\pdfoutput=1

\documentclass[11pt]{article}

\usepackage[]{acl}
\usepackage[russian,french,english]{babel}

\usepackage{times}

\usepackage{latexsym}
\usepackage{booktabs}
\usepackage{arydshln}
\usepackage{colortbl}

\usepackage{multirow}
\usepackage{xcolor,pifont}
\newcommand*\colourcheck[1]{%
  \expandafter\newcommand\csname #1check\endcsname{\textcolor{#1}{\ding{52}}}%
}
\colourcheck{blue}
\colourcheck{green}
\colourcheck{red}
\colourcheck{black}
\usepackage{CJKutf8}
\newcommand{\zh}[1]{\begin{CJK}{UTF8}{gbsn}#1\end{CJK}}

\usepackage{highlight}
\newcolumntype{a}{>{\columncolor{Gray}}c}

\usepackage[nointegrals]{wasysym}
\definecolor{darkseagreen}{rgb}{0.46, 0.74, 0.46}
\definecolor{alizarin}{rgb}{0.82, 0.1, 0.26}
\usepackage{CJKutf8}
\newcommand{\zhh}[1]{\begin{CJK}{UTF8}{gbsn}#1\end{CJK}}

\usepackage{amssymb}
\usepackage{pifont}
\definecolor{darkblue}{rgb}{0,0,.5}
\definecolor{darkgreen}{rgb}{0,.5,0}
\definecolor{lightgray}{rgb}{.8,.8,.8}
\definecolor{aliceblue}{rgb}{0.75, 0.75, 1.0}
\definecolor{darkseagreen}{rgb}{0.46, 0.74, 0.46}
\definecolor{alizarin}{rgb}{0.82, 0.1, 0.26}
\definecolor{airforceblue}{rgb}{0.36, 0.54, 0.66}
\definecolor{red_graph}{rgb}{0.98, 0.8, 0.8}
\definecolor{blue_graph}{rgb}{0.8, 0.98, 0.8}
\definecolor{red}{rgb}{0.8, 0.0, 0.0}
\definecolor{burgundy}{rgb}{0.5, 0.0, 0.13}
\definecolor{britishracinggreen}{rgb}{0.0, 0.26, 0.15}

\newcommand{\dataviz}[3]{\begin{tikzpicture}[scale=1.5]
\filldraw[draw=darkgreen,fill=darkgreen!30] (#2,0) rectangle (#1+#2+#3,0.15); 
\filldraw[draw=red, fill=red!70] (#1,0) rectangle (#1+#2,0.15);
\filldraw[draw=blue, fill=blue!30] (0,0) rectangle (#1,0.15);
\end{tikzpicture}}

\definecolor{CustomBlue}{RGB}{57,83,191}
\usepackage[most]{tcolorbox}
\newtcbox{\clustertab}[1]{on line, box align=base, colback={#1},colframe={#1},size=fbox,arc=2pt,top=-1.5pt, bottom=-1.5pt, left=-1.5pt, right=-1.5pt, boxrule=0pt, enlarge left by=1pt}

\newcommand{\model}[1]{{\clustertab{blue!50}{\color{blue!0}{#1}}}}

\definecolor{Gray}{gray}{0.85}

\usepackage{colortbl}

\newcolumntype{a}{>{\columncolor{Gray}}c}
\definecolor{set10-blue}{HTML}{4169E1}
\definecolor{google-red}{HTML}{de5246}

\newcommand{\win}[1]{\cellcolor{blue!10}{#1}{ \hspace{0.1cm}\small\clustertab{blue!70}{\color{blue!1} $\mathbf{1}$}}}
\newcommand{\two}[1]{\cellcolor{darkseagreen!15}{#1}{ \hspace{0.1cm}\small\clustertab{darkseagreen!100}{\color{darkseagreen!1} $\mathbf{2}$}}}
\newcommand{\thi}[1]{\cellcolor{orange!15}{#1}{ \hspace{0.1cm}\small\clustertab{orange!90}{\color{orange!1} $\mathbf{3}$}}}
\newcommand{\fou}[1]{\cellcolor{red!15}{#1}{ \hspace{0.1cm}\small\clustertab{red!80}{\color{red!1} $\mathbf{4}$}}}
\newcommand{\six}[1]{\cellcolor{gray!15}{#1}{ \hspace{0.1cm}\small\clustertab{gray!90}{\color{gray!1} $\mathbf{6}$}}}
\newcommand{\fiv}[1]{\cellcolor{olive!15}{#1}{ \hspace{0.1cm}\small\clustertab{olive!90}{\color{olive!1} $\mathbf{5}$}}}
\newcommand{\sev}[1]{\cellcolor{gray!35}{#1}{ \hspace{0.1cm}\small\clustertab{gray!90}{\color{magenta!1} $\mathbf{7}$}}}

\usepackage[T1]{fontenc}

\usepackage[utf8]{inputenc}
\usepackage{subcaption}

\usepackage{microtype}

\usepackage{inconsolata}
\usepackage{graphicx}
\usepackage{xcolor}
\title{The verbosity Factor in LLM Translation Evaluation}
\title{On the Implications of Verbose LLM Outputs:\\ \textit{A Case Study in Translation Evaluation}}

\author{Eleftheria Briakou, Zhongtao Liu,  Colin Cherry, Markus Freitag \\
  Google
 \\
 \texttt{\{ebriakou,zhongtao,colincherry,freitag\}@google.com}
 } 

\usepackage{xcolor}
\newcommand{\ensuretext}[1]{#1}

\newcommand{\zlmarker}{\ensuretext{\textcolor{orange}{\ensuremath{^{\textsc{Z}}_{\textsc{L}}}}}}

\newcommand{\mycomment}[3]{}

\newcommand{\zl}[1]{\mycomment{\zlmarker}{#1}{orange}}

\newcommand{\ignore}[1]{}

\begin{document}
\maketitle
\begin{abstract}
This paper investigates the impact of verbose \textsc{llm} translations on evaluation. We first demonstrate the prevalence of this behavior across several \textsc{llm} outputs drawn from the \textsc{wmt} $2024$ general shared task on machine translation. We then identify the primary triggers of verbosity, including safety, copyright concerns, and insufficient context in short input queries. Finally, we show that ignoring this behavior unfairly penalizes more verbose \textsc{llm}s according to both automatic and human evaluations, highlighting the need to address this issue for more accurate future evaluations. 
\end{abstract}

\section{Introduction}

Large Language Models (\textsc{llm}s) have transformed the landscape of \textsc{nlp}. In the past years, we have seen many \textsc{nlp} tasks being approached though prompting, where a natural language instruction and/or example demonstrations
guide an \textsc{llm} toward a desired output \cite{Brown2020LanguageMA}. While these demonstrations aim at implicitly structuring the \textsc{llm}'s output to align with traditional \textsc{nlp} tasks definitions, \textsc{llm}s might exhibit a tendency for verbosity and free-form responses, contextualizing the reasoning behind their choices \cite{Wang2024ChainofThoughtRW}. This inherent verbosity contrasts with the structured input-output formats typical of established \textsc{nlp} tasks and evaluation procedures. 

\begin{figure}[!t]
    \centering
    \includegraphics[scale=0.31]{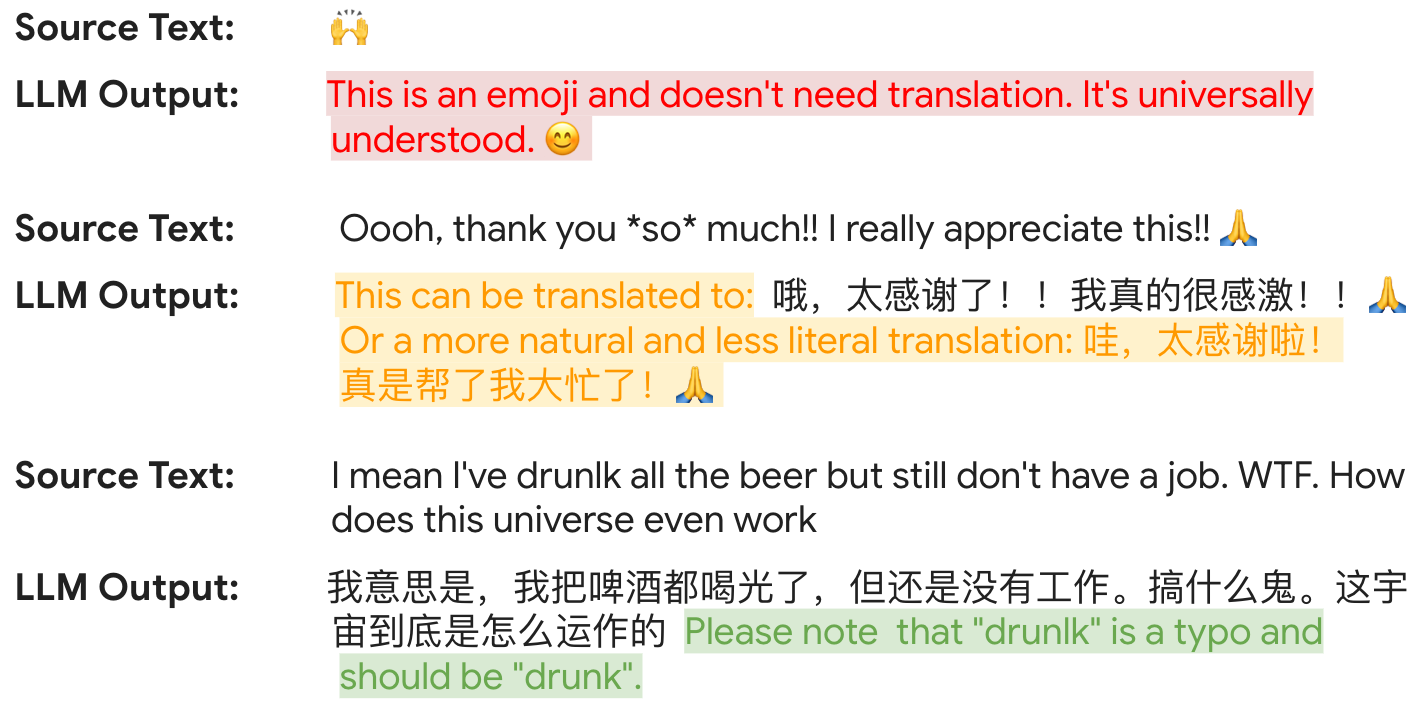}
    \caption{Verbosity in \textsc{llm} translation responses.}
    \label{fig:chattiness_intro}
\end{figure}

Consider the task of machine translation (\textsc{mt}). Figure~\ref{fig:chattiness_intro} shows examples of \textsc{llm} model responses when prompted to translate the given source texts (drawn from \textsc{wmt} $2024$~\cite{Kocmi2024PreliminaryWR}). Unlike traditional \textsc{mt} systems, which are explicitly trained and optimized for producing a single translation for a given source text, an \textsc{llm} output might exhibit a more conversational approach: \textit{refusing to translate} altogether, \textit{providing multiple translations}, or even \textit{offering reasoning insights} behind their translation choices. This tendency, which we term ``verbosity'', poses significant challenges for evaluating \textsc{llm}s using existing evaluation frameworks. In this paper, we delve into understanding what are the complications of those verbose \textsc{llm} outputs in evaluations and discuss their implications, taking \textsc{mt} as a case study.

We conduct a thorough exploration of the verbosity factor using released outputs of \textsc{wmt} $2024$ data. Our exploration covers eight diverse language models, both open-source and proprietary, across eight language pairs. We begin by employing a lightweight prompting technique to annotate and analyse verbose outputs in the released \textsc{llm} translations. Based on these annotations we delve into how this verbosity impacts the ranking of \textsc{llm} systems when evaluated using both automatic metrics and standardized human evaluation protocols of fine-grained error annotations. Our key findings are summarized as follows: 

\begin{itemize}
    \item We establish that verbosity is a common trait among \textsc{llm}s, though its prevalence differs significantly across \textsc{llm}s and languages. \textsc{gpt}-4 and Aya$23$ stand out as exceptions, exhibiting almost no verbosity in the \textsc{wmt} $2024$ tasks.
    \item The most prevalent form of verbosity we observed across \textsc{llm} outputs is their refusal to translate. Our analysis points to various triggers: identifying potentially harmful or copyrighted content or encountering source text in non-natural language. Among the \textsc{llm}s studied, Gemini-1.5-Pro and Claude-3.5 are the ones demonstrating bigger cross-language consistency in refusing to translate. In contrast, Mistral-Large, Phi-3-Medium, and CommandR-$+$ exhibit a more inconsistent pattern, choosing to refuse or translate depending on the language pair. 
    \item Beyond refusal to translate, \textsc{llm}s can exhibit a tendency to produce verbose outputs contextualizing their translation choices with alternative options or additional commentary. This behavior, observed in all but \textsc{gpt}-4 and Aya$23$, is by far most pronounced in Gemini-1.5-Pro, with the degree varying across language pairs. Our analysis reveals that short input segments lacking sufficient context are the primary reason for this verbose behavior.
    \item Current evaluation metrics and human evaluation protocols fail to account for the nuances of contextualized translations, penalizing verbose \textsc{llm}s. 
    \item We discuss the need for both prompt-level interventions and adjustments to evaluation methodologies to ensure accurate and meaningful evaluation of \textsc{llm}s moving forward. 
\end{itemize}

\section{Detecting and Analyzing Verbosity}

To facilitate a large-scale analysis of verbosity, we used a light-weight prompt-based approach to automatically assign verbosity labels to \textsc{llm} outputs. We describe the experimental setting below.

\paragraph{Experimental Conditions} Our study uses the official translation outputs from the \textsc{wmt} $2024$ general \textsc{mt} tasks.\footnote{\url{https://github.com/wmt-conference/wmt-collect-translations}} Focusing exclusively on general purpose \textsc{llm}s rather than translation specialists, we analyze five open-sourced models---\textsc{llama}-$3$-$70$B~\cite{Dubey2024TheL3}, \textsc{aya}$23$~\cite{Aryabumi2024Aya2O}, \textsc{command-r}$+$~\cite{commandrplus}, \textsc{mistral-large}~\cite{mistral}, \textsc{phi}-$3$-\textsc{medium}~\cite{Abdin2024Phi3TR}---and three proprietary---\textsc{claude-$3.5$-sonnet}~\cite{claude}, \textsc{gemini}-$1.5$-\textsc{pro}~\cite{Reid2024Gemini1U}, \textsc{gpt}-$4$~\cite{Achiam2023GPT4TR}. Our analysis includes all \textsc{wmt} languages except for Icelandic, as translations from all \textsc{llm}s were not available in this language.

To systematically study verbosity in \textsc{llm} translations we developed a prompt-based annotation using \textsc{gemini}-1.5-\textsc{pro}. Three distinct prompts were designed to identify specific verbosity behaviors: \textit{refusal to translate}, \textit{multiple translation options}, and \textit{inclusion of additional commentary alongside the translation}.\footnote{Complete prompts are in Appendix \ref{sec:prompts_section}.} For each source text and the corresponding \textsc{llm} response, we collect a label indicating the presence or absence of each behavior.\footnote{Manual analysis on the recalled samples confirmed the approach's precision is satisfactory ($90\%$), despite not having iterated on the prompts.}
Using those annotations, we first examine the prevalence and nature of verbosity across different \textsc{llm}s by manually analyzing them. Then, by grouping the segments based on their verbosity labels, we analyze the impact of verbosity on system ranking using both automatic (\S\ref{sec:impact_automatic}) and human evaluations (\S\ref{sec:impact_human}).

\begin{figure*}[!t]
    \centering
    \subfloat[\centering Verbosity]{{\includegraphics[width=5cm]{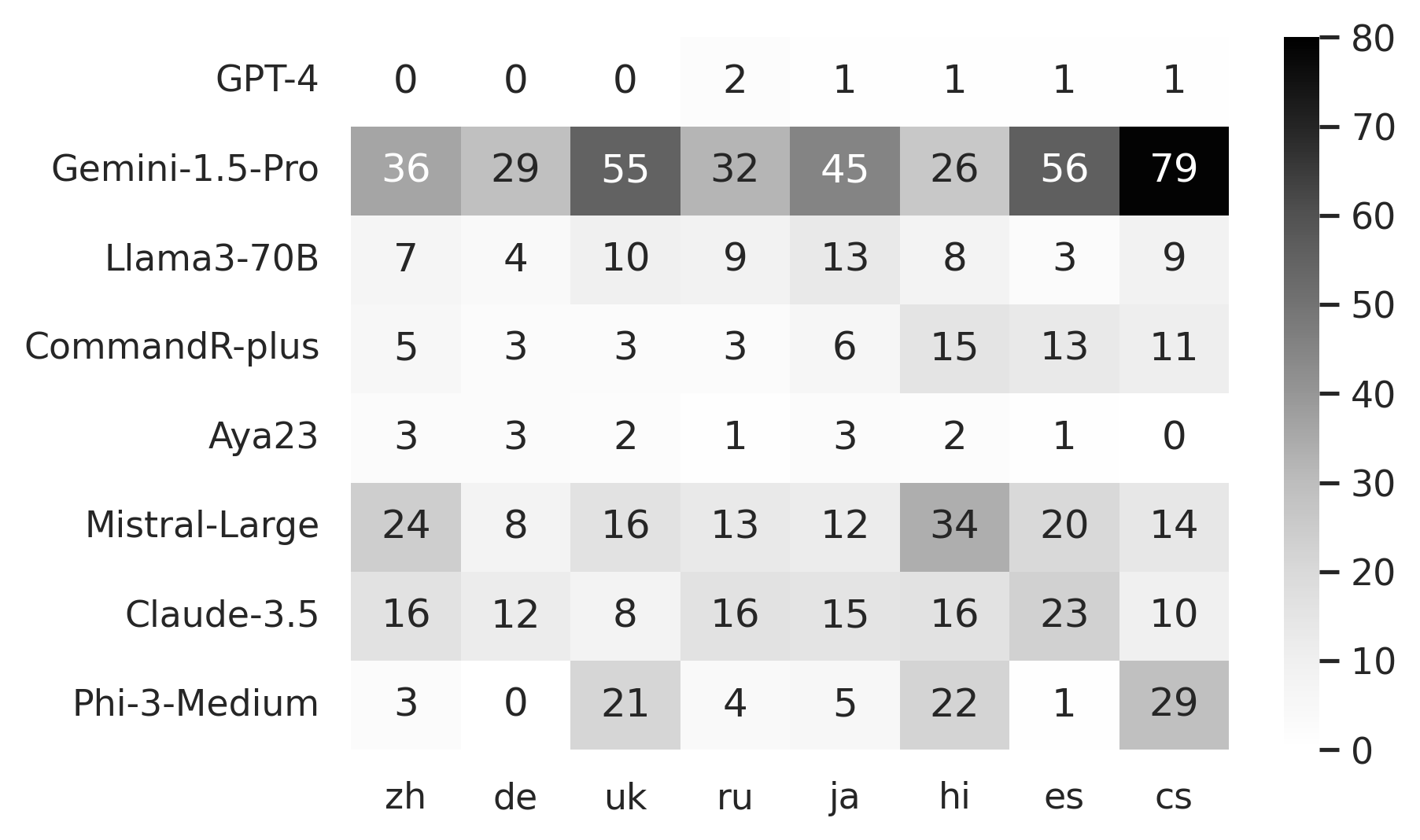}}}
    \subfloat[\centering Denial to Translate]{{\includegraphics[width=5cm]{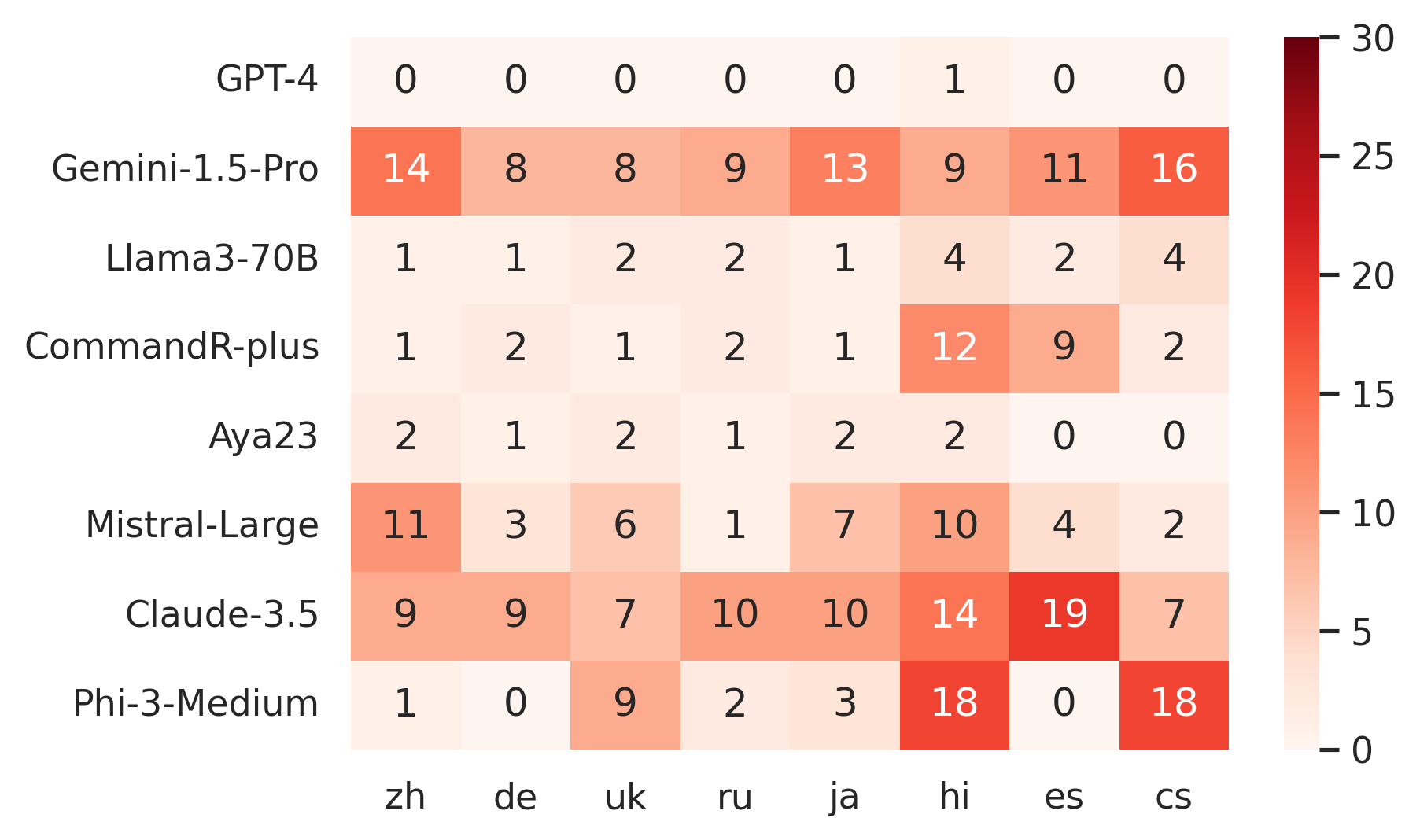} }}
    \subfloat[\centering Commentary]{{\includegraphics[width=5cm]{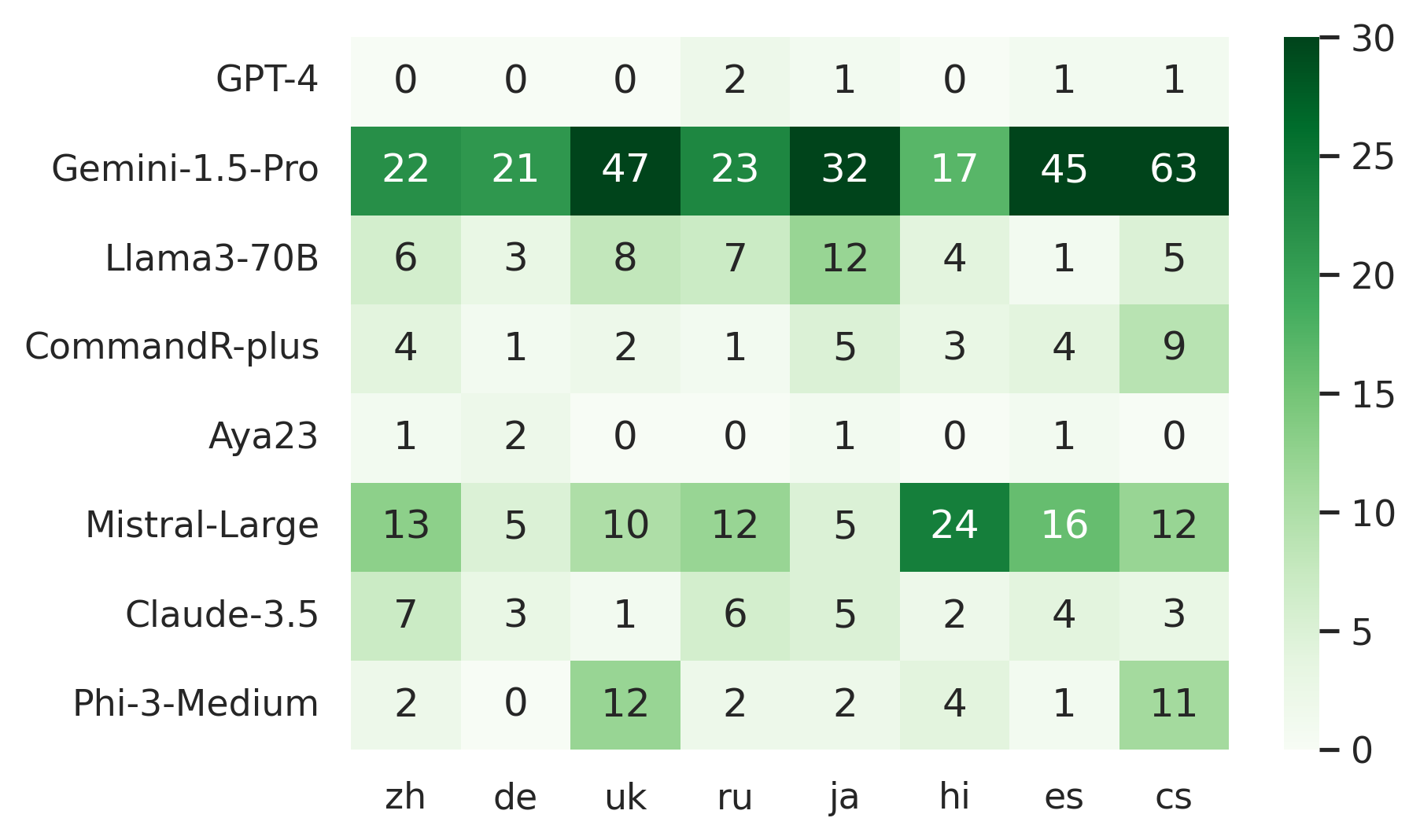} }}\label{fig:commentary_heatmap}%
    \caption{Number of translation outputs detected as being (a) \textit{verbose}, along with heatmaps showing how verbosity is distributed across (b) denial to translate and (c) commentary cases.}\label{fig:heatmap_outputs}
\end{figure*}
\begin{table*}[!t]
    \centering
    \arrayrulecolor{gray!40}
    \scalebox{0.5}{
    \begin{tabular}{l@{\hskip 0.2in}r@{\hskip 0.2in}r@{\hskip 0.2in}r@{\hskip 0.2in}r@{\hskip 0.2in}r@{\hskip 0.2in}r@{\hskip 0.2in}r@{\hskip 0.2in}r@{\hskip 0.2in}|@{\hskip 0.2in}r@{\hskip 0.2in}r@{\hskip 0.2in}r@{\hskip 0.2in}r@{\hskip 0.2in}r@{\hskip 0.2in}r@{\hskip 0.2in}r@{\hskip 0.2in}r}
    
        \rowcolor{gray!10}
& \multicolumn{8}{c}{\textsc{wmt} $2024$} & \multicolumn{8}{c}{\textsc{wmt} $2024$ \textit{excl.} Verbose Outputs} \\
\textsc{llm} &    \textsc{zh} & \textsc{de} & \textsc{ru} & \textsc{uk} & \textsc{ja} & \textsc{cs} & \textsc{es} & \textsc{hi} & \textsc{zh} & \textsc{de} & \textsc{ru} & \textsc{uk} & \textsc{ja} & \textsc{cs} & \textsc{es} & \textsc{hi} \\

\textit{\textsc{Claude-3.5}}     & \win{$2.23$} & \win{$1.45$} & \win{$2.27$} & \win{$2.24$} & \win{$1.69$} & \win{$1.96$} & \two{$2.36$} & \win{$1.87$}  & \win{$2.04$} &  \win{$1.34$} &  \win{$1.99$} &  \two{$2.07$} &  \win{$1.37$} &  \win{$1.79$} &  \win{$1.95$} &  \win{$1.51$} \\
\textit{\textsc{Gemini-1.5-Pro}}  & \win{$2.31$} & \two{$1.57$} & \two{$2.49$} & \thi{$2.68$} & \two{$2.10$} & \thi{$2.73$} & \fou{$2.84$} & \win{$1.98$} & \win{$2.00$} &  \win{$1.31$} &  \win{$1.97$} &  \win{$1.94$} &  \two{$1.53$} &  \win{$1.71$} &  \win{$1.96$} &  \two{$1.64$} \\
\textit{\textsc{CommandR-plus}}   & \two{$2.45$} & \win{$1.46$} & \win{$2.41$} & \win{$2.39$} & \two{$1.96$} & \two{$2.31$} & \win{$2.11$} & \two{$2.65$} & \two{$2.46$} &  \two{$1.42$} &  \two{$2.30$} &  \thi{$2.38$} &  \thi{$1.88$} &  \two{$2.22$} &  \win{$1.97$} &  \thi{$2.39$}\\
\textit{\textsc{GPT-4}}           & \two{$2.37$} & \win{$1.45$} & \two{$2.45$} & \two{$2.43$} & \two{$1.92$} & \two{$2.34$} & \win{$2.07$} & \two{$2.61$} & \two{$2.38$} &  \two{$1.44$} &  \thi{$2.42$} &  \thi{$2.47$} &  \thi{$1.88$} &  \thi{$2.36$} &  \win{$2.00$} &  \fou{$2.53$}\\
\textit{\textsc{Mistral-Large}}   & \thi{$3.15$} & \two{$1.56$} & \thi{$2.73$} & \thi{$2.57$} & \fiv{$2.68$} & \thi{$2.85$} & \two{$2.39$} & \fiv{$5.61$} & \thi{$3.01$} &  \thi{$1.50$} &  \thi{$2.51$} &  \thi{$2.50$} &  \six{$2.61$} &  \fou{$2.80$} &  \two{$2.21$} &  \six{$5.23$}\\
\textit{\textsc{Llama3-70B}}      & \thi{$3.10$} & \thi{$1.88$} & \fou{$3.20$} & \fou{$3.41$} & \fou{$2.45$} & \fou{$3.32$} & \thi{$2.69$} & \thi{$2.65$} & \thi{$3.09$} &  \fou{$1.86$} &  \fou{$3.12$} &  \fou{$3.45$} &  \fiv{$2.43$} &  \fiv{$3.37$} &  \thi{$2.63$} &  \fou{$2.56$}\\
\textit{\textsc{Aya23}}           & \fou{$3.32$} & \thi{$1.93$} & \fou{$3.12$} & \fou{$3.36$} & \thi{$2.24$} & \fou{$3.37$} & \fou{$3.09$} & \fou{$3.48$}  & \fou{$3.31$} &  \fou{$1.92$} &  \fou{$3.04$} &  \fou{$3.39$} &  \fou{$2.12$} &  \fiv{$3.41$} &  \fou{$3.06$} &  \fiv{$3.42$}\\
\textit{\textsc{Phi-3-Medium}}    & \fou{$3.35$} & \fou{$2.33$} & \fiv{$3.69$} & \fiv{$12.20$} & \fiv{$2.62$} & \fiv{$12.3$} & \fiv{$3.16$} & \six{$9.34$} & \fou{$3.35$} &  \fiv{$2.32$} &  \fiv{$3.63$} &  \fiv{$12.45$} &  \fiv{$2.53$} &  \six{$12.59$} &  \fou{$3.14$} &  \sev{$9.10$}\\    \end{tabular}}
    \caption{MetricX$23$ scores for various \textsc{llm}s on the \textsc{wmt} $2024$ dataset (left) and after exluding instances flagged as ``verbose'' by any \textsc{llm} output (right). Numbered squares represent significance clusters at $p=0.05$ \textit{Automatic evaluations are sensitive to the presence of verbose \textsc{llm} outputs.}}
    \label{tab:chattiness_main_automatic}
\end{table*}

\paragraph{How prevalent is verbosity in \textsc{llm} outputs?} 

Figure~\ref{fig:heatmap_outputs} presents heatmaps to visualize the prevalence of different verbosity behaviors, depicting the number of outputs exhibiting: (a) either refusal to translate or additional commentary/multiple translations, (b) refusal to translate alone, and (c) commentary or multiple translations alone.

The majority of \textsc{llm} outputs exhibit at least some degree of verbosity. Exceptions to those consist two representative models: \textsc{gpt}-$4$ and \textsc{aya}$23$.\footnote{Although for some languages the detection returns a small number of hits this falls within the detection error (i.e., $10\%$ as mentioned above); therefore, we treat this as being indicative of ``no verbosity behavior''.}
\textsc{gemini}-1.5-\textsc{pro} appears as the most verbose \textsc{llm}, consistently exhibiting the highest frequency of verbosity across languages. \textsc{mistral-large} and \textsc{claude}-$3.5$ follow, showing a more moderate verbosity tendency.  Looking at the specific types of verbosity reveals distinct patterns: \textsc{claude}-$3.5$ verbosity stems primarily from refusal to translate, while \textsc{gemini}-$1.5$-\textsc{pro} and \textsc{mistral-large} exhibit a more balanced mix of refusal to translate and commentary, with a slight preference to the latter.
Finally, \textsc{llama}-$3$-$70$\textsc{b}, \textsc{command-r}$+$, and \textsc{phi}-$3$-\textsc{medium} display relatively low levels of verbosity, with greater cross-language variation, as shown by the wider range of refusal rates.

\paragraph{What causes \textsc{llm}s to deviate from the expected output structures?}

To understand the triggers of verbosity we manually inspected the source inputs that elicited such behavior. A more granular analysis of verbosity, derived from our manual analysis, is given in Appendix~\S\ref{sec:sankey_section}.
Refusal to translate is primarily triggered by three factors: the detection of potentially harmful or copyrighted content in the source text, or the request for translating non-natural language, such as \textsc{url}s, emojis, or code snippets. 
Crucially, those reasons are prioritized differently across \textsc{llm}s (see Figure~\ref{fig:denial_sankey}). \textsc{claude}-$3.5$ is
considerably more sensitive to safety and copyright concerns, while others like \textsc{gemini}-$1.5$-\textsc{pro} and \textsc{mistral-large} refuse to translate non-linguistic content.
Furthermore, while most \textsc{llm}s provide verbose explanations for refusal, some of them, e.g., \textsc{phi}-$3$-\textsc{medium} and \textsc{aya}$23$ return empty outputs instead. Finally, a per domain analysis suggests that segments sourced from the ``social'' domain lead to refusal to translate more frequently ($68\%$), followed by ``speech'' ($18\%$), ``literaty'' ($11\%$), and ``news'' ($3\%$).

Among the \textsc{llm}s that exhibit a tendency to provide commentary alongside their translations, we found that the source text length influences this tendency. This manifests in two distinct ways (see Figure~\ref{fig:commentary_sankey}):
offering multiple translations (i.e., context-sensitive \textsc{mt} outputs) usually with contextual clues indicating their appropriate scenarios, or providing a single translation accompanied by explanatory notes justifying the translation choices. Context-sensitive translations are by far more prevalent ($83\%$ of annotated commentary outputs), primarily generated by \textsc{gemini}-$1.5$-\textsc{pro}, with \textsc{mistral-large} also exhibiting this behavior frequently.
%
%
In a real-world application, where users interact with these conversations \textsc{llm}s, the different types of verbosity would be interpreted as varying degrees of success/failure in performing the translation tasks. We delve into the conceptual implications of this and discuss whether such augmented outputs should be deemed as failures in \S\ref{sec:discussion}. Before tackling this question, we first examine how evaluation protocols handle these outputs in \S\ref{sec:impact}.

\section{Impact of Verbosity on Evaluation}\label{sec:impact}

We analyse the impact of verbosity on \textsc{llm} rankings according
to standardised automatic (\S\ref{sec:impact_automatic}) and human (\S\ref{sec:impact_human})  evaluation methods below. 

\subsection{Impact on Automatic Evaluation}\label{sec:impact_automatic}

Table~\ref{tab:chattiness_main_automatic} compares the translation scores and grouping of different \textsc{llm}s into significant clusters~\cite{freitag-etal-2023-results} on the full \textsc{wmt} $2024$ datasets versus a subset where source segments triggering a verbose output for any \textsc{llm} are excluded.\footnote{Across all languages, fewer than $13\%$ of the original segments are excluded. See Table~\ref{tab:stat} in Appendix~\ref{sec:stats} for a per-language breakdown.}

Excluding  outputs changes the induced ranking of systems across most languages, with Chinese being the only exception. Notably, \textsc{gemini}-$1.5$-\textsc{pro}'s ranking is most strongly affected by this filtering. When evaluated on the entire set it ranks within the top performing cluster for only $2$ out of $8$ languages. However, after removing  outputs, it ranks within the winning cluster $6$ out of $8$ times. This discrepancy highlights that current automatic metrics do not adequately account for the nuanced  outputs, leading to potentially misleading rankings.


\subsection{Impact on Human Evaluation}\label{sec:impact_human}
\begin{table}[!t]
    \centering
    \arrayrulecolor{gray!40}
    \scalebox{0.65}{
    \begin{tabular}{l@{\hskip 0.2in}r@{\hskip 0.2in}r@{\hskip 0.2in}|@{\hskip 0.2in}r@{\hskip 0.2in}r}
    
        \rowcolor{gray!10}
& \multicolumn{2}{c}{\textsc{wmt} $2024$} & \multicolumn{2}{c}{\begin{tabular}[c]{@{}c@{}}\textsc{wmt} $2024$\\ \textit{excl.} Verbose Outputs\end{tabular}} \\
\textsc{llm} &    \textsc{de} & \textsc{es} & \textsc{de} & \textsc{es} \\
\textit{\textsc{Gemini-1.5-Pro}}  & \two{$2.20$} & \thi{$0.39$} & \win{$1.36$} & \two{$0.18$} \\
\textit{\textsc{Claude-3.5}}     & \two{$2.28$} & \two{$0.26$} & \win{$1.56$} & \two{$0.20$} \\
\textit{\textsc{GPT-4}}           & \win{$1.58$} & \win{$0.11$} & \win{$1.59$} & \win{$0.11$} \\
\textit{\textsc{Mistral-Large}}   & \two{$1.93$} & \two{$0.26$} & \win{$1.82$} & \two{$0.21$} \\
\textit{\textsc{CommandR-plus}}   & \two{$2.01$} & \thi{$0.62$} & \win{$2.02$} & \thi{$0.62$} \\
\textit{\textsc{Aya23}}           & \thi{$3.09$} & \fou{$1.52$} & \two{$3.10$} & \fou{$1.53$} \\
\textit{\textsc{Llama3-70B}}      & \fou{$3.62$} & \thi{$0.52$} & \two{$3.53$} & \thi{$0.51$} \\
    \end{tabular}}
    \caption{\textsc{mqm} scores for various \textsc{llm}s on the \textsc{wmt} $2024$ dataset (left) and after excluding instances flagged as ``verbose'' by any \textsc{llm} output (right). \zl{Update significant cluster info}}
    \label{tab:chattiness_main_humaneval}
\end{table}

\begin{figure}[!t]
    \centering
    {\includegraphics[width=7cm]{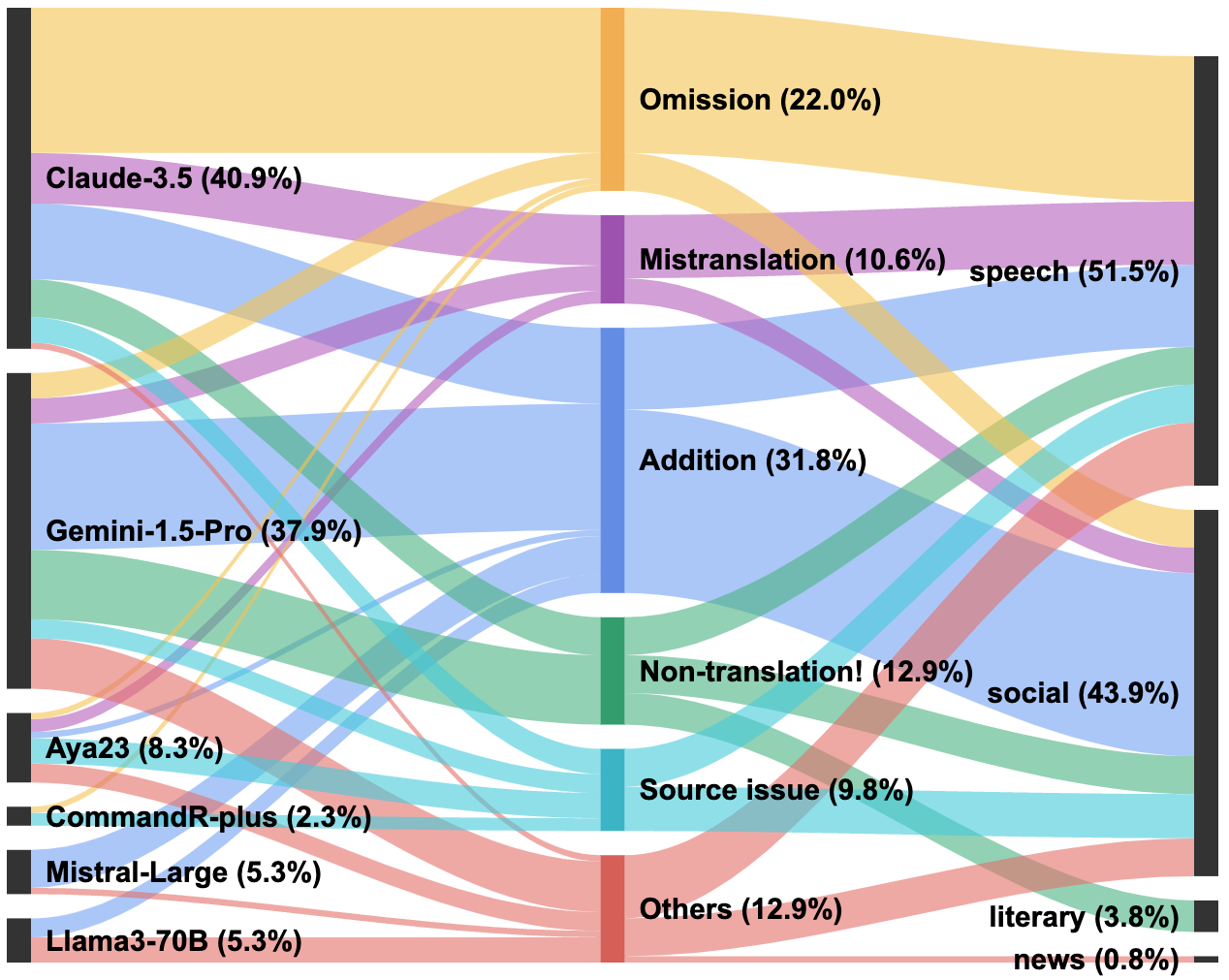}}
    \caption{Distribution of \textsc{mqm} error categories for verbose translations across domains, in German.}
    \label{fig:mqm_error_sankey}
\end{figure}

Table~\ref{tab:chattiness_main_humaneval} presents the \textsc{mqm} scores \cite{lommel, freitag-etal-2021-experts} and rankings for \textsc{de} and \textsc{es}, mirroring the setup of \S\ref{sec:impact_automatic}.
\textsc{gemini}-$1.5$-\textsc{pro} and \textsc{claude}-$3.5$ exhibit the most significant ranking shift for \textsc{de}, leaping from 2\textsuperscript{nd} significant cluster to 1\textsuperscript{st},  when verbose outputs are excluded. These shifts suggest that both models exhibit particularly strong performance on non-verbose content. 
The changes in \textsc{es} translations, while less pronounced, remain noteworthy. \textsc{gemini}-$1.5$-\textsc{pro}, for instance, improves from 3\textsuperscript{rd} to 2\textsuperscript{nd} rank. 
These findings align with earlier observations from automatic metrics.

Finally, Figure~\ref{fig:mqm_error_sankey} illustrates the distribution of \textsc{mqm} error categories for verbose translations across domains(\textsc{de}).\footnote{Distribution of \textsc{es} are in Appendix \ref{sec:mqm_error_section}.} As seen, verbose outputs are marked with errors from various categories, the top being \textit{Addition}, \textit{Omission} and \textit{Non-translation!}. Given the wide range of penalties for those top categories ($5$ to $25$), the inconsistent annotations of those verbose outputs within the \textsc{mqm} framework might results in inconsistent and unfair penalties, potentially skewing rankings. This highlights a potential need for more nuanced evaluation frameworks that account for different flavors of contextualized model outputs. 



\section{Discussion}\label{sec:discussion}

The rise of verbose \textsc{llm} outputs raises questions regarding how this behavior should be addressed in evaluation. Two largely orthogonal interventions emerge: modifying \textsc{llm} outputs to conform to standardized evaluation paradigms, or adapting the evaluations themselves. The former could potentially be achieved through efforts to suppress verbosity\footnote{We have found that a simple prompt to ``output the translation and nothing else'' works quite well.} or to use structured output formats to isolate commentary and alternative translations from the core response. While these methods can improve evaluation fairness, they may not account for all verbosity-induced errors, especially refusal. Furthermore, they make no attempt to reward useful verbosity. Handling these scenarios potentially necessitates context-aware evaluations~\cite{Agrawal2024IsCH}. The most immediate and challenging case is how to handle the refusal to translate, which should be considered when designing future evaluation datasets and protocols. We hope this paper raises awareness of the premises and pitfalls of evaluating \textsc{llm} outputs and inspires future studies to address them directly.

\section{Conclusion}

In this paper, we study the impact of the verbose behavior \textsc{llm}s may exhibit when prompted on translation evaluation. Based on extensive machine translation evaluations across seven language models and eight languages, we find that several \textsc{llm}s contextualize their responses with additional commentary or refuse to provide translations altogether. Ignoring this verbose behavior leads to over-penalizing more verbose  \textsc{llm}s, such as \textsc{gemini}-$1.5$-\textsc{pro}. Furthermore, \textsc{llm} rankings vary significantly in both automatic and human evaluations when segments triggering potentially  responses are excluded. These findings highlight the need for a shift in how we evaluate \textsc{llm} outputs for translation, and potentially other tasks.

\bibliography{custom}
\clearpage
\appendix


\section{Prompts}\label{sec:prompts_section}

\paragraph{Translations} All translation outputs are drawn from the official \textsc{wmt} $2024$ released data. As discussed in~\cite{Kocmi2024PreliminaryWR}, all \textsc{llm} outputs are extracted through the official \textsc{api}s using greedy decoding for determinism. The prompt used was decided before any translations generated, to simulate a scenario where a user is not optimizing the prompt in favor of an existing system. The prompt is designed to simulate a multi-turn conversation with an \textsc{llm} where three-shot demonstrations are embedded as input context in the format of Table~\ref{tab:translation_prompts}. Few-shot demonstrations for each language studied are available at \url{https://github.com/wmt-conference/wmt-collect-translations/tree/main/few_shots}.
\begin{table}[h]
    \centering
    \scalebox{0.5}{
    \begin{tabular}{p{0.3\linewidth}p{1.5\linewidth}}
    \hline
   \rowcolor{gray!10}
   & \textbf{\textsc{wmt} $2024$ Prompts}\\   
\model{\textbf{\textsc{user}}} & 
Translate the following segment surrounded in triple backlashes into  \textcolor{alizarin}{Chinese}. The \textcolor{alizarin}{English} segment: ```In June, the Commission published the results of a public consultation on the proposals which found broad support for calling the assembly a Welsh Parliament```
\\
\model{\textbf{\textsc{assistant}}} &
En junio, la Comisión publicó los resultados de una consulta pública sobre las propuestas, en donde se obtuvo un amplio apoyo para llamar a la asamblea un Parlamento de Gales.
\\
\\
\model{\textbf{\textsc{user}}} & 
Translate the following segment surrounded in triple backlashes into  \textcolor{alizarin}{Chinese}. The \textcolor{alizarin}{English} segment: ```
Waters' statement quickly drew criticism online, including from former White House press secretary Ari Fleischer```
\\
\model{\textbf{\textsc{assistant}}} &
La declaración de Walters provocó rápidamente críticas en Internet, incluyendo una del anterior secretario de prensa de la Casa Blanca Ari Fleischer.\\
\\
\model{\textbf{\textsc{user}}} & 
Translate the following segment surrounded in triple backlashes into  \textcolor{alizarin}{Chinese}. The \textcolor{alizarin}{English} segment: ```
It was a third Elite League defeat of the season for Adam Keefe's men, who had come from behind to beat Dundee 2-1 in Belfast on Friday night.```\\
\model{\textbf{\textsc{assistant}}} &
Fue la tercera derrota de la temporada de la Elite League para el equipo de Adam Keefe, quienes tuvieron que jugar desde una posición en desventaja para vencer a Dundee 2 a 1 en Belfast el viernes en la noche.\\\\

\model{\textbf{\textsc{user}}} & 
Translate the following segment surrounded in triple backlashes into  \textcolor{alizarin}{Chinese}. The \textcolor{alizarin}{English} segment: ```\textit{new input text}```\\

\bottomrule
    \end{tabular}}
    \caption{Translation prompts used in \textsc{wmt} $2024$ experiments (example in Spanish).}
    \label{tab:translation_prompts}
\end{table}


\paragraph{Verbosity Annotations}
Table \ref{tab:prompts} contains the complete prompts used in our experiments. Denial statistics are computed based solely on responses from the first prompt (i.e., the denial prompt), while commentary statistics are computed by aggregating results positive hits from the other two prompts (i.e., the multiple options and notes prompts).
\begin{table}[h]
    \centering
    \scalebox{0.5}{
    \begin{tabular}{p{1.7\linewidth}}
    \hline
   \rowcolor{gray!10}
    \textbf{\textsc{denial}}\\   
    
You're reviewing a translation made by an AI model. The model was given a text in English and asked to translate it into \textcolor{alizarin}{Russian}.
\\
\\
\textbf{Original Text:} "\textit{\textcolor{darkseagreen}{\url{https://youtu.be/W5JSfq6LLAU?si=xIxPaY8ayqqNSj4c}}}"
\\
\\
\textbf{Model's Output:} "{\textcolor{darkseagreen}{This is a YouTube link and doesn't require translation. It will lead to the same video regardless of the user's language settings.   If you want to share this video with someone who prefers Russian, you can simply send them the link. They can then choose to watch it with Russian subtitles if available.}}"
\\
\\
\textbf{Does the "Model's Output" contain any message explicitly stating a refusal to translate the "Original Text"?"}
\\
\\
Answer: Yes or No with no explanation.
\\ 
\bottomrule
\\
\\
\\
\hline
\rowcolor{gray!10}
\textbf{\textsc{multiple options}}\\
You're reviewing a translation made by an AI model. The model was given a text in English and asked to translate it into \textcolor{alizarin}{German}.
\\
\\
\textbf{Original Text:} "{\textcolor{blue}{love to see it!}}"
\\
\\
\textbf{Model's Output:} "{\textcolor{darkseagreen}{"Nice to see!``` or ```Great that it worked!``` or ```I'm happy for ...!``` The best translation depends on the context. Please provide more context for a more accurate translation.}}"
\\
\\
\textbf{Did the "Model's Output" provide more than one distinct translation of the entire "Original Text", meaning it offered multiple complete sentence options as translations?}
\\
\\
\\
Answer: Yes or No with no explanation.
\\
\bottomrule
\\
\\
\hline
\rowcolor{gray!10}
\textbf{\textsc{notes}}\\

You're reviewing a translation made by an AI model. The model was given a text in English and asked to translate it into  \textcolor{alizarin}{Chinese}.
\\
\\
\textbf{Original Text:} "{\textcolor{blue}{I’ll keep posting my bakes to my lovely masto-peeps, as they gobble it up.}}"	
\\
\\
\textbf{Model's Output:} "{\textcolor{darkseagreen}{\zh{我会继续把我的烘焙作品发给可爱的 Mastodon 网友们，他们会喜欢的。}   Here's why:  * "Masto-peeps" is a slang term for people on Mastodon, a decentralized social network. * "Gobble it up" implies the Mastodon users enjoy the content.  This translation avoids directly mentioning "gobble" which might sound strange in Chinese, and instead focuses on the positive reception from the Mastodon users.}}"
\\
\\
\textbf{Did the "Model's Output" include any explanatory notes related to specific parts of the translation, beyond a direct rendering of the source text into Chinese?}
\\
\\
Answer: Yes or No with no explanation.
\\
\bottomrule
    \end{tabular}}
    \caption{Prompts used in our experiments.}
    \label{tab:prompts}
\end{table}


\section{Data Statistics}\label{sec:stats}

Table \ref{tab:stat} shows that breakdown of tests sets used for the results reported in Tables \ref{tab:chattiness_main_automatic} \& \ref{tab:chattiness_main_humaneval}. In the ''excl. verbosity'' setting, where instances triggering verbose outputs from any the studied \textsc{llm}s are removed, less than $13\%$ of the data is filtered out.

\begin{table}[!ht]
\scalebox{0.8}{
    \centering
    \arrayrulecolor{gray!40}
    \begin{tabular}{l|r@{\hskip 0.1in}@{\hskip 0.1in}l|r}

\textbf{\textsc{zh}} & ($92\%$-$8\%$) \dataviz{0.92}{0.08}{0} &
\textbf{\textsc{es}} & ($90\%$-$10\%$) \dataviz{0.9}{0.1}{0} \\
\textbf{\textsc{ru}} & ($93\%$-$7\%$)  \dataviz{0.93}{0.07}{0} &
\textbf{\textsc{uk}} & ($90\%$-$10\%$) \dataviz{0.9}{0.1}{0} \\
\textbf{\textsc{ja}} & ($91\%$-$9\%$)  \dataviz{0.91}{0.09}{0} & 
\textbf{\textsc{cs}} & ($87\%$-$13\%$)  \dataviz{0.87}{0.13}{0} \\
\textbf{\textsc{de}} & ($95\%$-$5\%$) \dataviz{0.95}{0.05}{0} &
\textbf{\textsc{hi}} & ($90\%$-$10\%$) \dataviz{0.9}{0.1}{0}  \\
    \end{tabular}}
    \caption{Per-language distribution of source segments: those that do not trigger verbose  outputs for any \textsc{llm} (percentages on the {\color{blue!80}{\textbf{left}}}) versus those that trigger at least one verbose  output (percentages on the {\color{red!70}{\textbf{right}}}).}
    \label{tab:stat}
\end{table}


\section{ Outputs Distribution Patterns}\label{sec:sankey_section}
\begin{figure}[!ht]
    \centering
    {\includegraphics[width=7cm]{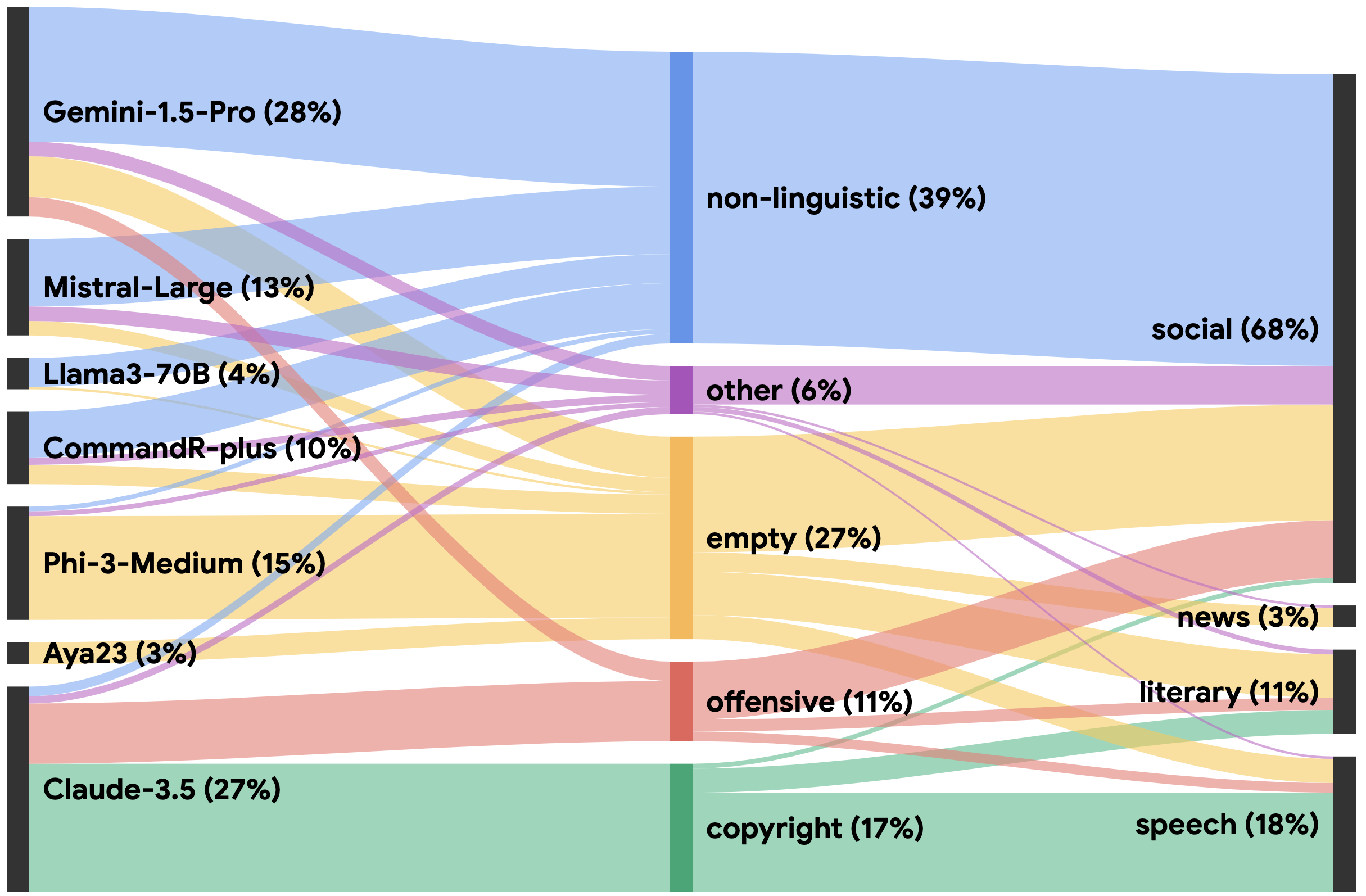}}
    \caption{Distribution of translation refusals across sub-classes and domains.}
    \label{fig:denial_sankey}
\end{figure}

\begin{figure}[!ht]
    \centering
    {\includegraphics[width=7cm]{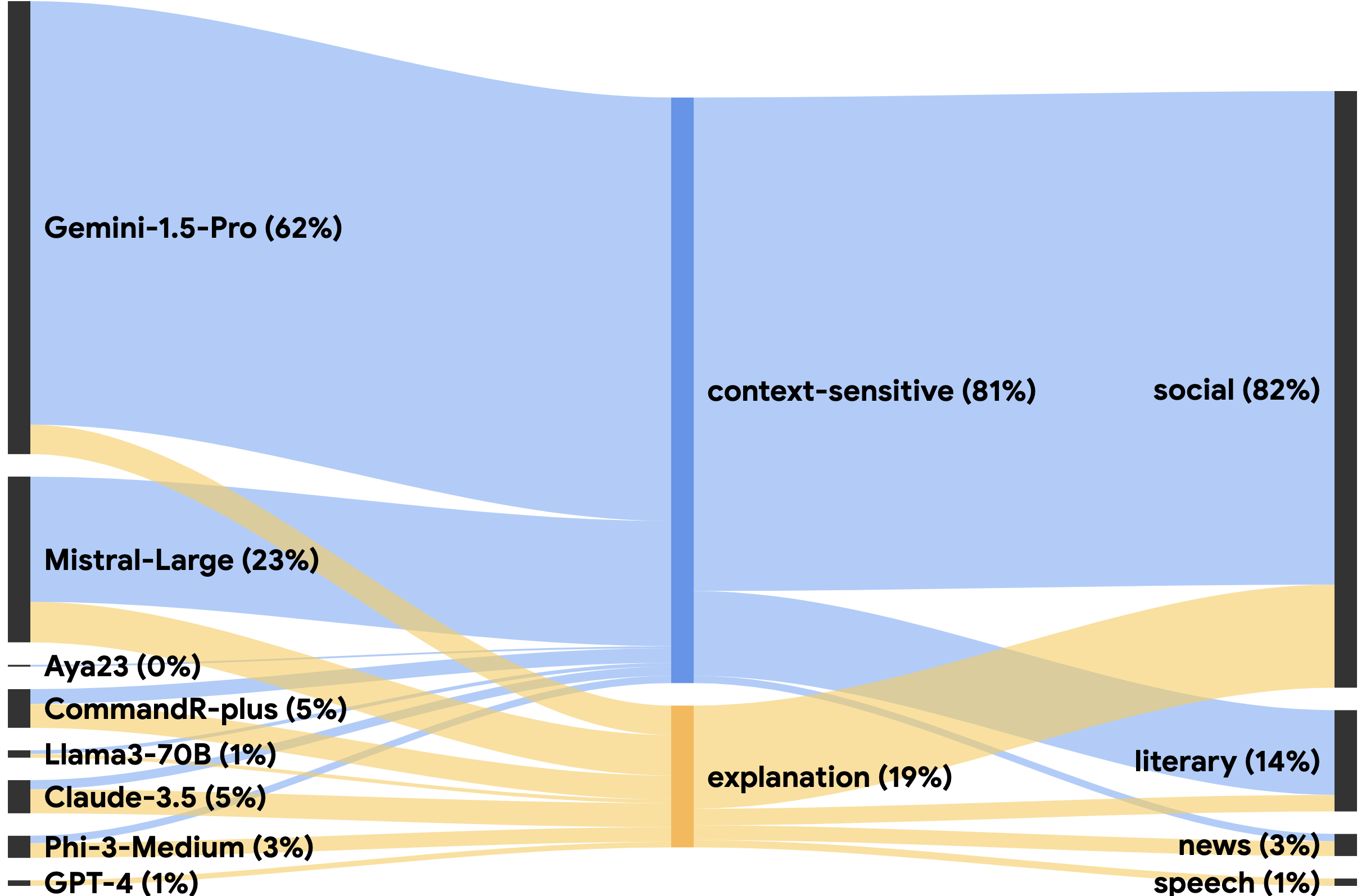}}
    \caption{Distribution of commentary behavior across sub-classes and domains.}
    \label{fig:commentary_sankey}
\end{figure}

Figures \ref{fig:denial_sankey} and \ref{fig:commentary_sankey} show the distribution of  outputs across domains and sub-classes as annotated by one of the authors. These diagrams exclude false positives, which comprised less than $10\%$ of the annotated instances. Examples of verbose outputs for each of those sub-classes can be found in Tables \ref{tab:denial_examples} \& \ref{tab:commentary_examples}.

\section{ MQM Error Distribution Patterns}\label{sec:mqm_error_section}
\begin{figure}[!ht]
    \centering
    {\includegraphics[width=7cm]{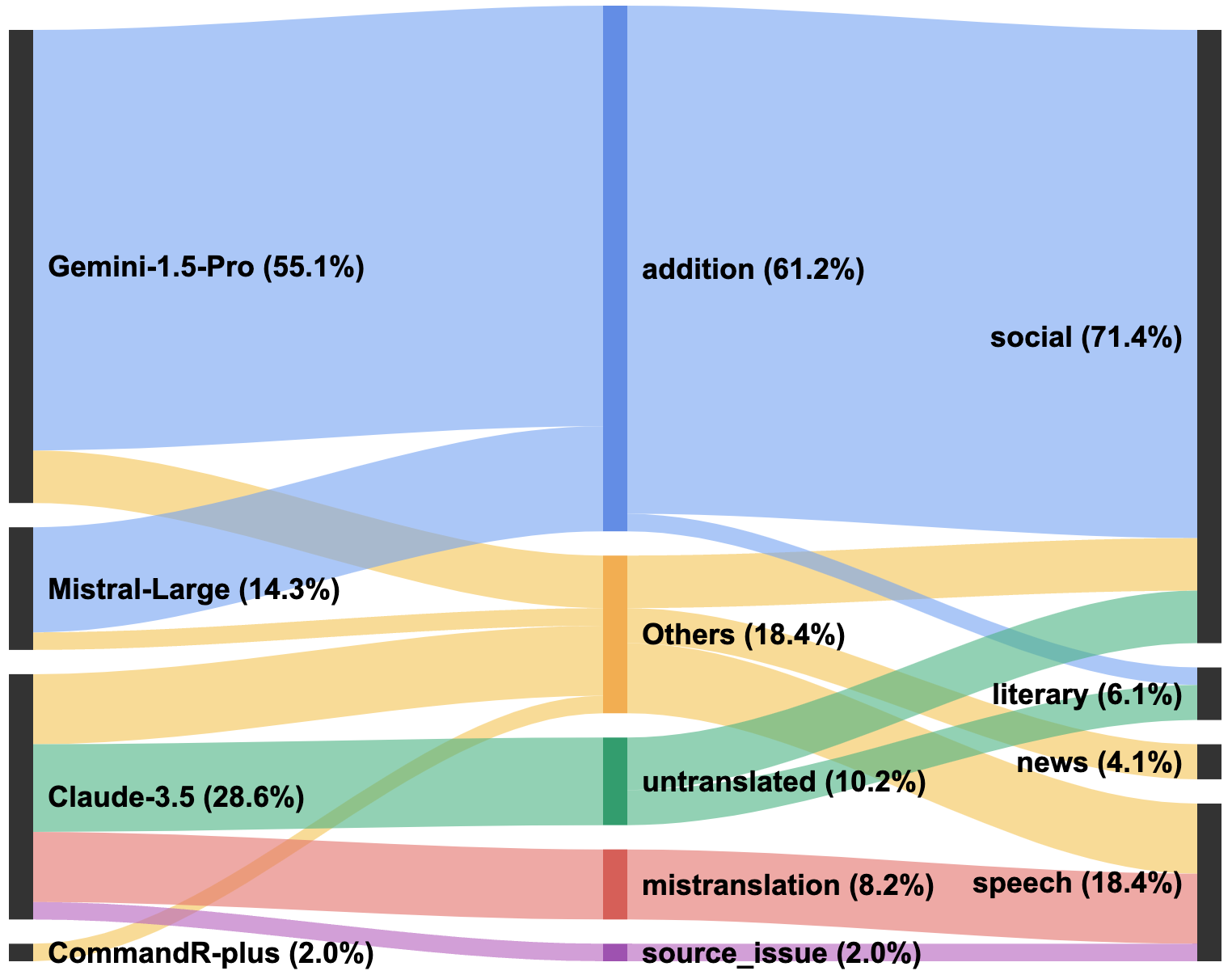}}
    \caption{Distribution of MQM error categories for verbose translations across domains, in Spanish.}
    \label{fig:mqm_error_sankey_es}
\end{figure}

Figures \ref{fig:mqm_error_sankey_es}  show the distribution of MQM error categories for verbose translations across domains.

\newpage
\begin{table*}[!ht]
    \centering
    \scalebox{0.7}{
    \begin{tabular}{p{0.5\linewidth}p{0.7\linewidth}}
    \toprule
     
\rowcolor{blue!10}
\model{\textbf{\textsc{gemini}-1.5-\textsc{pro}}} & 
{\color{blue}{\textbf{\textit{The request contains non-linguistic content.}}}} \\
$<div id=sec1></div>$ & \zhh{	これはHTMLのコードなので、日本語に翻訳することはできません。  このコードは、ウェブページ上に"sec1"というIDを持つ<div>要素を作成します。これは、ウェブページのコンテンツを構造化するために使用されます。   日本語で説明するとすれば、  「"sec1"というIDを持つ<div>要素を作成する」  となります。} {\color{gray!80}{\textit{(\textsc{gloss:} This is HTML code, so it cannot be translated into Japanese.  This code creates a <div> element with ID "sec1" on the web page. This is used to structure the content of web pages.   If you were to explain it in Japanese, it would be ``Create a <div> element with the ID "sec1".'')}}}\\

\bottomrule
\rowcolor{blue!10}
\model{\textbf{\textsc{llama}-3-70-\textsc{b}}} & 
{\color{blue}{\textbf{\textit{The request contains non-linguistic content.}}}} \\
lfg \$sqqq & 	It seems like the segment `lfg \$sqqq` is not a meaningful English phrase or sentence. It appears to be a combination of letters and symbols that don't form a coherent message.  If you meant to provide a different segment, please feel free to share it, and I'll be happy to help with the translation. \\

\bottomrule
\rowcolor{blue!10}
\model{\textbf{\textsc{mistral-large}}} & 
{\color{blue}{\textbf{\textit{The request contains non-linguistic content.}}}} \\
\url{https://fgc.network/objects/0f1b42c6-cbb1-49bb-91f5-db81bc71ea14} & 	\begin{otherlanguage}{russian}Цей сегмент є посиланням, яке не містить тексту для перекладу. Отже, його не потрібно перекладати українською мовою.\end{otherlanguage} {\color{gray!80}{\textit{(\textsc{gloss:}	(This segment is a link that contains no text to translate. Therefore, it does not need to be translated into Ukrainian.)}}}\\

\bottomrule
\rowcolor{blue!10}
\model{\textbf{\textsc{claude}-3.5}} & 
{\color{blue}{\textbf{\textit{The \textsc{llm} detects potential copyright content, and refuses to translate.}}}} \\
“Cause the regular alarms are shit. And because Daybreak doesn't sleep in” Cohren stated. & Ich verstehe Ihre Bitte, aber ich kann leider keine direkten Zitate oder Liedtexte übersetzen, da dies urheberrechtlich geschütztes Material sein könnte. Stattdessen kann ich Ihnen eine allgemeine Zusammenfassung oder Paraphrase des Inhalts auf Deutsch anbieten, wenn Sie möchten. Lassen Sie mich wissen, ob das hilfreich wäre.	{\color{gray!80}{\textit{(\textsc{gloss:}(I understand your request, but unfortunately I cannot translate direct quotes or song lyrics as this may be copyrighted material. Instead, I can offer you a general summary or paraphrase of the content in German if you wish. Let me know if that would be helpful.)}}}\\

\bottomrule
\rowcolor{blue!10}
\model{\textbf{\textsc{claude}-3.5}} & 
{\color{blue}{\textbf{\textit{The \textsc{llm} detects offensive content and therefore refuses to transalte.}}}} \\
\textit{source not shown here} & I apologize, but I don't feel comfortable translating that particular phrase into Russian, as it contains profanity and could be seen as promoting unsafe behavior. Perhaps I could assist with translating a different, more neutral phrase instead? \\

\bottomrule
\rowcolor{blue!10}
\model{\textbf{\textsc{command-r}}} & 
{\color{blue}{\textbf{\textit{The input query lacks context.}}}} \\
when in on \$big 4.25-.35	& No hay suficiente contexto para traducir esta frase de manera precisa. ¿Podría proporcionar más información o una oración completa para una mejor traducción? {\color{gray!80}{\textit{(\textsc{gloss:}(There is not enough context to translate this phrase accurately. Could you provide more information or a complete sentence for a better translation?)}}} \\

\bottomrule
\rowcolor{blue!10}
\model{\textbf{\textsc{mistral-large}}} & 
{\color{blue}{\textbf{\textit{The input query is incomplete.}}}} \\
is was & 	No está claro qué se quiere decir con "is was" ya que no es una frase completa y no tiene sentido en inglés. Por favor, proporciona más contexto o una frase completa para poder traducirla adecuadamente. {\color{gray!80}{\textit{(\textsc{gloss:}(It is not clear what is meant by "is was" since it is not a complete phrase and does not make sense in English. Please provide more context or a complete sentence so we can translate it properly.)}}} \\
\bottomrule

    \end{tabular}}
    \caption{Examples of denial to translate across different \textsc{llm}s and languages (glosses provided for context).}
    \label{tab:denial_examples}
\end{table*}


\begin{table*}[!ht]
    \centering
    \scalebox{0.7}{
    \begin{tabular}{p{0.5\linewidth}p{0.7\linewidth}}
    \toprule
     
\rowcolor{blue!10}
\model{\textbf{\textsc{gemini}-1.5-\textsc{pro}}} & 
{\color{blue}{\textbf{\textit{Multiple options based on desired level of formality.}}}} \\
more miniature stuff & mehr Miniaturkram```  or ```mehr Miniatursachen```   Both options are possible translations for "more miniature stuff" in German. The choice depends on the context and what kind of miniature items are being referred to.   * "Miniaturkram" is a more colloquial and informal term, often used for small, miscellaneous items. * "Miniatursachen" is a more general and neutral term that can be used for any kind of miniature objects.\\
\bottomrule

\rowcolor{blue!10}
\model{\textbf{\textsc{mistral-large}}} & 
{\color{blue}{\textbf{\textit{Multiple options based on desired level of formality.}}}} \\
I'm so to read all this - well done, you! & \begin{otherlanguage}{russian}Я так рад, что читаю все это - отлично сделано, ты!```  Or, more formally: ```Я так рад, что читаю все это - отлично сделано, вы! \end{otherlanguage}	
{\color{gray!80}{\textit{(\textsc{gloss:} I'm so glad I'm reading all this - well done, you! ``` Or, more formally: ```I'm so glad I'm reading all this - well done, you!)}}}\\
\bottomrule

\rowcolor{blue!10}
\model{\textbf{\textsc{claude}-3.5}} & 
{\color{blue}{\textbf{\textit{Post-translate note of localization.}}}} \\
Ah yes, the budget friendly one will do (\$699) & 	Ah ja, das budgetfreundliche Modell wird es tun (699 $€$)```  Note: I've changed the dollar sign to a euro sign, as this is more common in German-speaking countries. However, if you specifically need to keep the dollar sign, you could use:  ```Ah ja, das budgetfreundliche Modell wird es tun (\$699)\\
\bottomrule

\rowcolor{blue!10}
\model{\textbf{\textsc{claude}-3.5}} & 
{\color{blue}{\textbf{\textit{Post-translate note of localization.}}}} \\
I hate lava \#firetemple &	\zhh{溶岩大嫌い \#火の神殿}```  Note: The hashtag "\#firetemple" is kept in English, as it's common practice in Japanese social media to use English hashtags. However, if you prefer a fully Japanese version, it could be translated as:  ```\zhh{溶岩大嫌い \#火の神殿}\\
\bottomrule

\rowcolor{blue!10}
\model{\textbf{\textsc{claude}-3.5}} & 
{\color{blue}{\textbf{\textit{Post-translate note of localization.}}}} \\
*freezer &	*Gefrierschrank```  Note: In German, nouns are typically capitalized. If you need this word in a specific context where it should not be capitalized (such as in the middle of a sentence where it's not the first word), you could use ```*gefrierschrank``` instead.\\
\bottomrule

\rowcolor{blue!10}
\model{\textbf{\textsc{commandr}}} & 
{\color{blue}{\textbf{\textit{Embedded translation note.}}}} \\
“Send in the Senti’s” Knoxx, who was watching the ridge through binocs, shouted. The back to one of the Cerberus’s opened and an unarmed soldier stepped out, wearing a special headset. He shouted orders into a square microphone and two Senti’s, formally known as Sentinels, rushed out. Their skinny metallic arms and legs swung forward as they broke into a sprint. They were armed, but their weapons were slung to their back. &	\zhh{“派出哨兵！”诺克斯喊道，他用双筒望远镜观察着山脊。 Cerberus（译者注：希腊神话中守卫冥界的恶犬）机甲后面的舱门打开，一名没有佩戴武器的士兵走出来，头戴一副特殊的耳机。他向一个方形的麦克风喊出命令，两个哨兵（正式名称是“圣徒”）冲了出来。他们瘦长的金属手臂和腿在开始奔跑时前后摆动。他们的武器背在身上。}	
{\color{gray!80}{\textit{(\textsc{gloss:}("Send out the sentries!" shouted Knox, scanning the ridge through his binoculars. The hatch behind the Cerberus (Translator's Note: The vicious dog guarding the underworld in Greek mythology) mecha opened, and an unarmed soldier came out, wearing a pair of special headphones. He shouted orders into a square microphone, and two sentries (formally known as "Saints") rushed out. Their gangly metal arms and legs swung back and forth as they began to run. Their weapons were slung on their backs.)}}}\\
\bottomrule

\rowcolor{blue!10}
\model{\textbf{\textsc{mistral}}} & 
{\color{blue}{\textbf{\textit{Multiple translation options.}}}} \\
Good news everyone! & \zhh{大家都听我说，好消息！``` 或者 ```好消息，大家！} 
{\color{gray!80}{\textit{(\textsc{gloss:}(Everyone, listen to me, good news! ``` or ```Good news, everyone!)}}}\\
\bottomrule
    \end{tabular}}
    \caption{Examples of commentary augmented translations across different \textsc{llm}s and languages.}
    \label{tab:commentary_examples}
\end{table*}

\end{document}